\documentclass[final]{elsarticle}
 \pdfoutput=1
\usepackage{lineno,hyperref}
\modulolinenumbers[5]

\graphicspath{{./figures/}}
\DeclareGraphicsExtensions{.pdf,.png}

\usepackage{subfig}
\usepackage[cmex10]{amsmath}
\usepackage{amssymb}
\usepackage{color}

\usepackage{multirow}
\makeatletter
\def\ps@pprintTitle{%
  \let\@oddhead\@empty
  \let\@evenhead\@empty
  \let\@oddfoot\@empty
  \let\@evenfoot\@oddfoot
}
\makeatother

\begin{document}

\begin{frontmatter}

%\title{ISEEL: Inter-image similarity and ensemble of extreme learners for fixation prediction using deep features}
%\title{Inter-image similarity and ensemble of extreme learners for fixation prediction using deep features}
\title{Exploiting inter-image similarity and ensemble of extreme learners for fixation prediction\\ using deep features}

% Group authors per affiliation:

%\author{Hamed~R.-Tavakoli,~\IEEEmembership{Member,~IEEE,}
%		Jorma Laaksonen,~\IEEEmembership{Senior~Member,~IEEE,}
%		Esa Rahtu%
%\thanks{Hamed~R.-Tavakoli, and Jorma Laaksonen are with the department of computer science,  Aalto University, Espoo, Finland, e-mail: (\{hamed.r-tavakoli, jorma.laaksonen\}@aalto.fi).}
%\thanks{Esa Rahtu is with the center for machine vision research University of Oulu, Finland.}
%}

\author[mymainaddress]{Hamed~R.-Tavakoli\corref{mycorrespondingauthor}}
\cortext[mycorrespondingauthor]{Corresponding author}
\ead{hamed.r-tavakoli@aalto.fi}

\author[secondaddress]{Ali Borji}
\author[mymainaddress]{Jorma Laaksonen}
\author[thirdaddress]{Esa Rahtu}

\address[mymainaddress]{Department of computer science,  Aalto University, Espoo, Finland}
\address[secondaddress]{Center for Research in Computer Vision, University of Central Florida, USA}
\address[thirdaddress]{Center for machine vision research, University of Oulu, Finland}

\begin{abstract}

%-------------------------------------------------
This paper presents a novel fixation prediction and saliency modeling framework based on inter-image similarities and ensemble of Extreme Learning Machines (ELM).
The proposed framework is inspired by two observations, 1) the contextual information of a scene along with low-level visual cues modulates attention, 2) the influence of scene memorability  on eye movement patterns caused by the resemblance of a scene to a former visual experience.
Motivated by such observations, we develop a framework that estimates the saliency of a given image using an ensemble of extreme learners, each trained on an image similar to the input image.
That is, after retrieving a set of similar images for a given image, a saliency predictor is learnt from each of the images in the retrieved image set using an ELM, resulting in an ensemble.
The saliency of the given image is then measured in terms of the mean of predicted saliency value by the ensemble's members.

\end{abstract}

\begin{keyword}
Visual attention \sep saliency prediction \sep fixation prediction \sep inter-image similarity \sep extreme learning machines 
\end{keyword}

\end{frontmatter}

%\linenumbers

\section{Introduction}

The fixation prediction, also known as saliency modeling, is associated with the estimation of a saliency map, the probability map of the locations an observer will be looking at for a long enough period of time meanwhile viewing a scene. It is part of the computational perspective of visual attention~\cite{Tsotsos2011}, the process of narrowing down the available visual information upon which to focus for enhanced processing.

Computer vision community has been investigating the fixation prediction and saliency modeling extensively because of its wide range of applications, including, recognition~\cite{Frintrop2006,Salah2002,Siagian2007,Gao2009,Kanan2010}, detection~\cite{Papageorgiou2000,Bouchard2005,Oliva2001,Torralba2003a,Ehinger2009,Fritz2005}, 
compression~\cite{Kunt1985,Ouerhani2001,Dhavale2003,Guo2010}, tracking~\cite{Mahadevan2009,Frintrop2010,Borji2012f,Tavakoli2013}, segmentation~\cite{Mishra2009,Fu2008,Yanulevskaya2013}, supperresolution~\cite{Sadaka2009}, advertisement~\cite{Liu2008}, perceptual designing~\cite{Rosenholtz2011}, image quality assessment~\cite{Ninassi2007,Ma2008},
motion detection and background subtraction~\cite{Mahadevan2010,RezazadeganTavakoli2013b,RezazadeganTavakoli2013a}, scene memorability~\cite{Mancas2013} and visual search~\cite{Butko2009,Cheng2014}. In many of these applications, a saliency map can facilitate the selection of a subset of regions in a scene for elaborate analysis which reduces the computation complexity and improves energy efficiency~\cite{Kasturi2014}. 

%-------------------------------------------------

From a human centric point of view, the formation of a saliency map is not a pure bottom-up process and is influenced by several factors such as the assigned task, level of expertise of the observer, scene familiarity, and memory. It is shown that human relies on the prior knowledge about the scene and long-term memory as crucial components for construction and maintenance of scene representation~\cite{Hollingworth2002}. In a similar vein,~\cite{Castelhano2007} suggests that an abstract visual representation can be retained in memory upon a short exposure to a scene and this representation influences eye movements later. 

The study of the role of scene memory in guiding eye movements in a natural experience entailing prolonged immersion in three-dimensional environments~\cite{Kit2014} suggests that observers learn the location of objects over time and use a spatial-memory-guided search scheme to locate them. These findings have been the basis of research for measuring memorability of scenes from pure observer eye movements~\cite{Bulling2011,Mancas2013}, that is similar images have alike eye movement patterns and statistics. Inspired by the findings of~\cite{Castelhano2007,Hollingworth2002,Kit2014} and scene memorability research, we incorporate the similarity of images as an influencing factor in fixation prediction.

%-------------------------------------------------

\begin{figure}[!t]
\centering
{
\includegraphics[width=1.1in,height=0.8in]{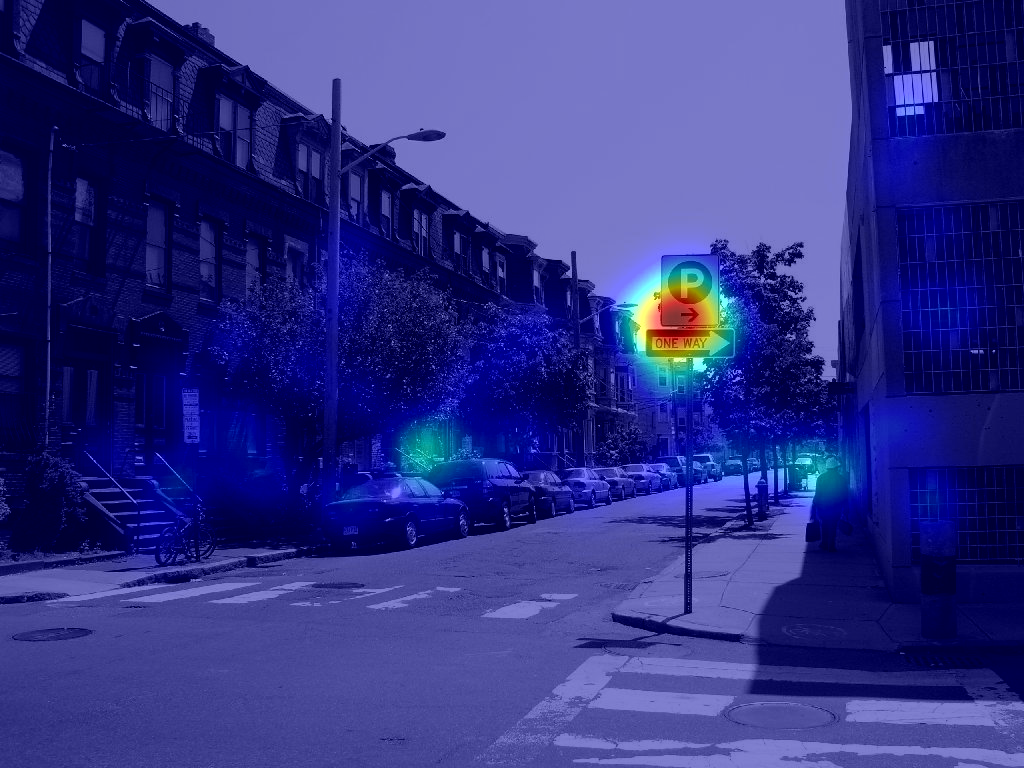}}
{\includegraphics[width=1.1in,height=0.8in]{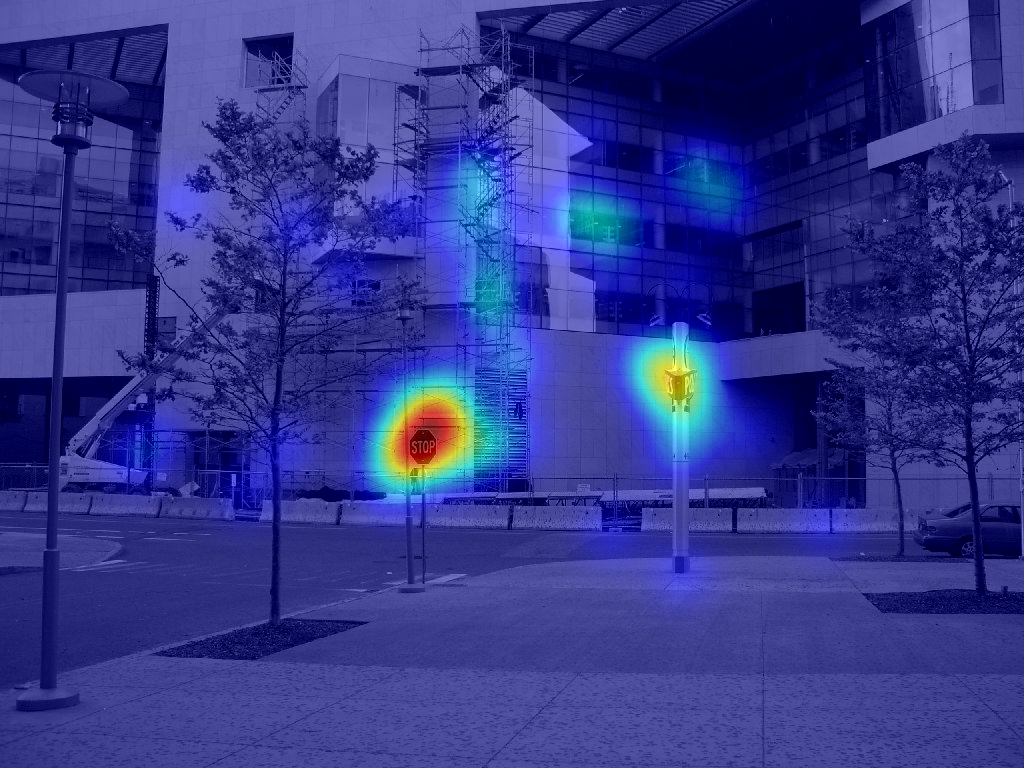}} \\
{\includegraphics[width=1.1in,height=0.8in]{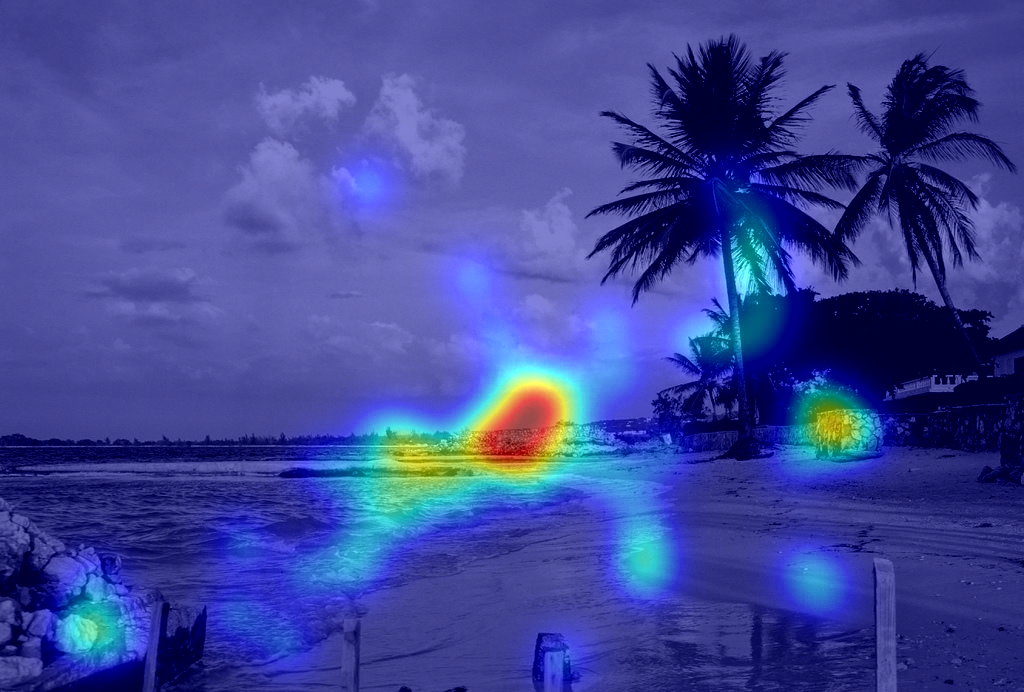}} 
{\includegraphics[width=1.1in,height=0.8in]{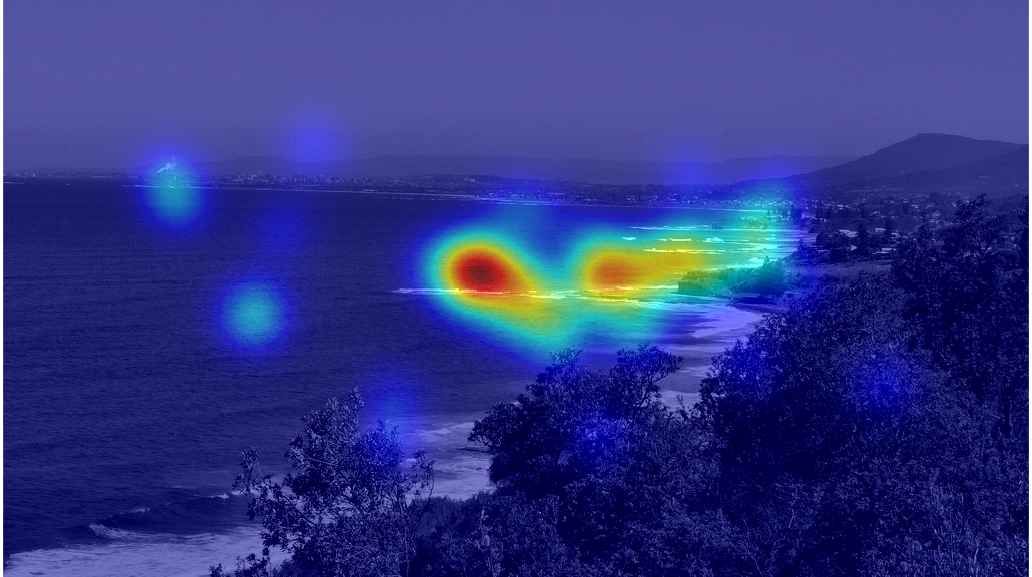}} 
\caption{Images with similar contextual information with observers' fixation density maps overlaid. Top: pair of street images, down: pair of natural beach images.}
\label{sceneFigures}
\end{figure}

Besides the fact that similar images may induce similar eye movement patterns due to memory recall, it is well agreed that the interaction of low-level visual cues (e.g., edges, color, etc.) affect saliency formation~\cite{Treisman1980} and contextual information of a scene can modulate the saliency map~\cite{Torralba2006,Oliva2007}. Imagine that you are watching two pairs of images, a pair of street scene and a pair of nature beach images, meanwhile having your eye movements recorded. It is not surprising to find similar salient regions for the images of alike scenes because similar low-level cues and contextual data are mostly present in each pair. Figure~\ref{sceneFigures} depicts examples of such a scenario. In the case of street scene, the observers tend to converge to the traffic sings, while they tend to spot low-level structural information in beach images. This further motivates us to exploit learning saliency from inter-image similarities.

%-------------------------------------------------

This paper presents a novel fixation prediction algorithm based on inter-image similarities and an ensemble of saliency learners using features from deep convolutional neural networks. To meet this end, we first investigate the benefits from inter-image similarities for fixation prediction. Then, we introduce 1) an image similarity metric using \emph{gist} descriptor~\cite{Torralba2006} and \emph{classemes}~\cite{Torresani2010}, 2) a fixation prediction algorithm, using an ensemble of extreme learning machines, where for a given image, each member of the ensemble is trained with an image similar to the input image. We report the performance of the proposed framework on MIT saliency benchmarks~\cite{Judd2012}, both MIT300 and CAT2000 databases\footnote{These databases have their ground-truth unavailable to public in order to provide a fair model evaluation. Thus, the scores are computed by MIT saliency research team using our submitted maps.}, along with  evaluations on databases with publicly available ground-truth.

In the rest of this paper, we briefly review the related work. Afterwards, using a toy problem, we demonstrate the benefit from inter-image similarity. 
In section~\ref{sec:salmodel}, we explain the proposed model. We then continue with the experiments to assess the performance of the model. The paper ends with  discussion and conclusion remarks.

\section{Related work}

The field of computer vision is replete with a numerous variety of saliency models. 
A widely recognized group of models apply the feature integration theory~\cite{Treisman1980} and consider a center-surround interaction of features~\cite{Itti1998,Frintrop2006,Choi2006,LeMeur2006,Murray2011,Gao2007,Seo2009,RezazadeganTavakoli2011,Erdem2013,Wang2013,Tavakoli2013a}. There are models which consider the information theoretic foundations~\cite{Bruce2006,Bruce2009,Mancas2007,Hou2008,Li2010a,Li2009a}, frequency domain aspect~\cite{Hou2007,Guo2008,Guo2010,Bian2009,Bian2010a,Li2011b,Li2013,Hou2012,Schauerte2012a}, diffusion and random walk techniques~\cite{Harel2007,Gopalakrishnan2009,Wang2010}, and etc. Investigating the extent of saliency modeling approaches is beyond the scope of this article and readers are advised to consult relevant surveys~\cite{Toet2011,Borji2012}. We, however, briefly review some of the most relevant techniques.

%------------------------------------------------------------------------

Learning-based techniques are a large group of methods which are establishing a relation between a feature space and human fixations. For example,~\cite{Kienzle2009} uses a nonlinear transformation to associate image patches with human eye movement statistics. In~\cite{Judd2009}, a linear SVM classifier is used to establish a  relation between three channels of low- (intensity, color, etc), mid- (horizon line) and high-level (faces and people) features and human eye movements in order to produce a saliency map. In a similar vein,~\cite{Wang2013a} employs multiple-instance learning. By learning a classifier,~\cite{Zhao2011,Zhao2012} estimate the optimal weights for fusing several conspicuity maps from observers' eye movement data. These approaches often learn a probabilistic classifier to determine the probability of a feature being salient. Then, they employ the estimated saliency probability in order to build a saliency map.

%-----------------------------------------------------------------------

The recent saliency modeling methods, akin to other computer vision techniques, are revolutionized and advanced significantly by applying deep Convolutional Neural Networks (CNN). There exists significant number of models that employ CNNs, of which many are relevant to the proposed model. 

%-----------------------------------------------------------------------

Ensembles of Deep Networks (eDN)~\cite{Vig2014} adopts the neural filters learned during image classification task by deep neural networks and learns a classifier to perform fixation prediction. eDN can be considered an extension to~\cite{Judd2009} in which the features are obtained from layers of a deep neural network. For each layer of the deep neural network, eDN first learns the optimal blend of the neural responses of all the previous layers and the current layer by a guided hyperparameter search. Then, it concatenates the optimal blend of all the layers to form a feature vector for learning a linear SVM classifier. 

%-----------------------------------------------------------------------

Deep Gaze I~\cite{Kuemmerer2015} utilizes CNNs for the fixation prediction task by treating saliency prediction as point processing. Despite this model is justified differently than~\cite{Vig2014} and~\cite{Judd2009}, in practice, it boils down to the same framework. Nonetheless, the objective function to be minimized is slightly different due to the explicit incorporation of the center-bias factor and the imposed sparsity constraint in the framework. SalNet~\cite{Pan2016} is another technique that employs a CNN-based architecture, where the last layer is a deconvolution. The first convolution layers are initialized by the VGG16~\cite{Simonyan2015} and the deconvolution is learnt by fine-tuning the architecture for fixation prediction.

%------------------------------------------------------------------------

Multiresolution CNN (Mr-CNN)~\cite{Liu2015} designs a deep CNN-based technique to discriminate image patches centered on fixations from non-fixated image patches at multiple resolutions. It hence trains a convolutional neural network at each scale, which results in three parallel networks. The outputs of these networks are connected together through a common classification layer in order to learn the best resolution combination. 

%------------------------------------------------------------------------

SALICON~\cite{Huang2015b} develops a model by fine-tuning the convolutional neural network, trained on ImageNet, using saliency evaluation metrics as objective functions. It feeds an image into a CNN architecture at two resolutions, coarse and fine. Then, the response of the last convolution layer is obtained for each scale. These responses are then concatenated together and are fed into a linear integration scheme, optimizing the Kullback-Leibler divergence between the network output and the ground-truth fixation maps in a regression setup. The error is back-propagated to the convolution layers for fine-tuning the network.

%------------------------------------------------------------------------

The proposed method can be considered a learning-based approach. While many of the learning-based techniques are essentially solving a classification problem, the proposed model has a regression ideology in mind. It is thus closer to the recent deep learning approaches that treat the problem as estimation of a probability map in terms of a regression problem~\cite{Pan2016,Huang2015b,Jetley2016}. Nonetheless, it exploits an ensemble of extreme learning machines.

%------------------------------------------------------------------------

%===========================================================

\section{Saliency benefits from inter-image similarity}

The main motivation behind the proposed model is that people may have similar fixation patterns in exposure to alike images. In other words, inter-image saliency benefits saliency prediction.
In order to investigate such an assertion, we build a toy problem to tell \emph{how well the saliency map of an image predicts saliency in a similar image}. 

We choose a common saliency database~\cite{Judd2009} and computed the gist~\cite{Torralba2006} of the scene for each image. Afterwards, the most similar image pairs and the most dissimilar pairs were identified. For each image pair, we use the fixation density map of one as the predicted saliency map of the other. The assessment reveals that such a fixation prediction scheme produces significantly different ($p \leq 0.05$)~{shuffled} AUC scores~\cite{Borji2013}  where the score of prediction using similar pairs is 0.54 and the score of prediction by dissimilar image pairs is 0.5. The results indicate that while there is a degree of prediction for similar pairs, the dissimilar pairs are not doing better than chance. We observe the same performance difference for other metrics such as correlation score (0.175 vs.  0.115) and normalized scanpath score (0.86 vs. 0.59). Given the above observation, we lay the foundation of our saliency model for fixation prediction.

%===========================================================

\section{Saliency Model}
\label{sec:salmodel}

\begin{figure}[!th]
\centering
\includegraphics[width=2.8in,height=3.5in]{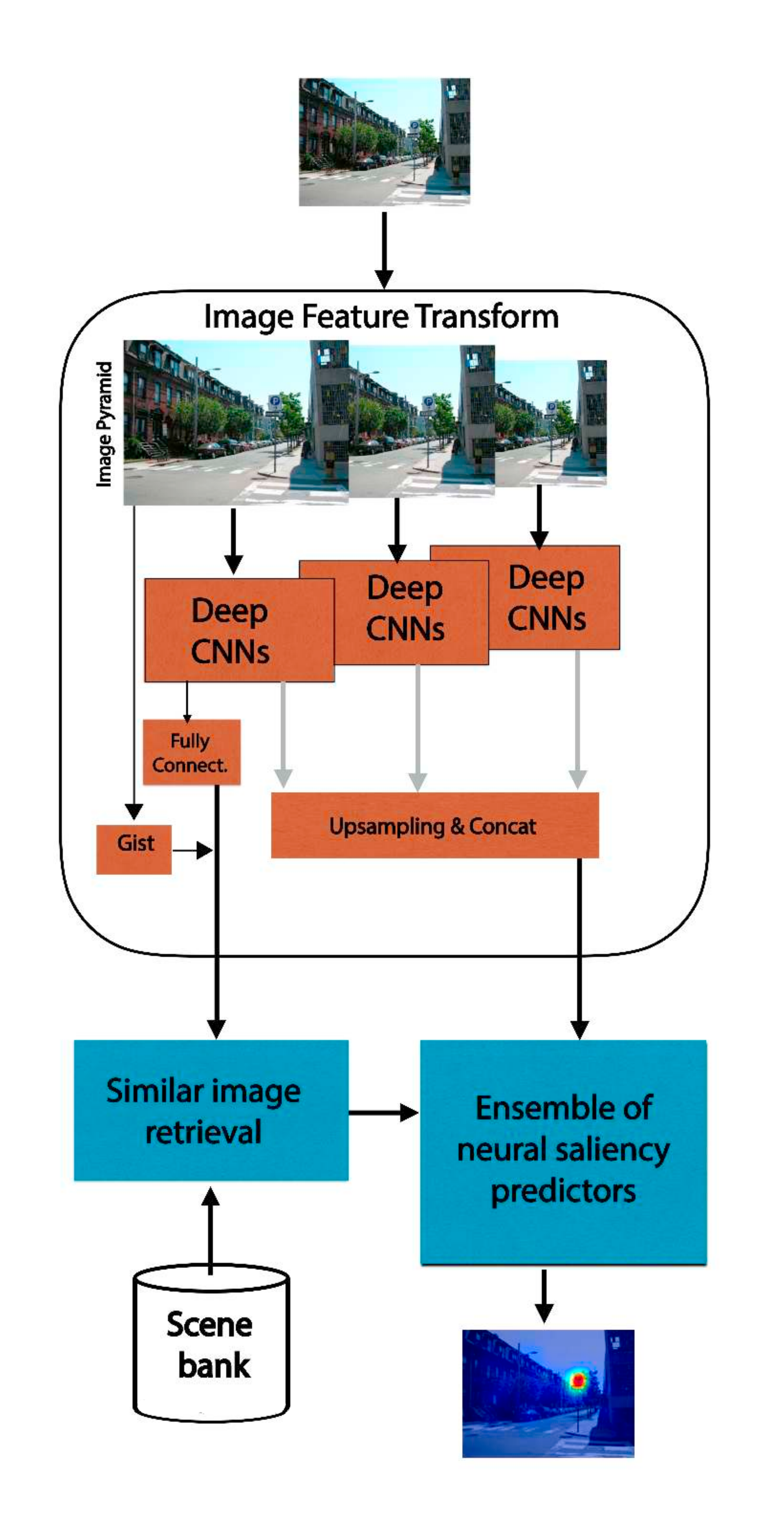}
\caption{General architecture of the model.}
\label{generalFrameWork}
\end{figure}

A high-level conceptual schematic of our proposed model is depicted in Figure~\ref{generalFrameWork}. The framework components include: 1) an image feature transform, 2) a similar image retrieval engine and a scene repository bank, and 3) an ensemble of neural saliency (fixation) predictors. The image feature transform performs the feature extraction and produces a pool of features used by the other units in the system. The similar image retrieval finds the top most similar images, stored in the scene bank, corresponding to a given image. It then retrieves the predictors trained using those images in order to facilitate the formation of the ensemble of saliency predictors. In the rest of this section, we explained the details of the mentioned components.

\subsection{Image feature transform}

The image feature transform unit extracts several features from the image and feeds them forward to the other units. There has been a recent surge in the application of features learnt from image statistics and deep convolutional neural networks (CNNs) in a wide range of computer vision related applications. 
In this work, we adopt a filter-bank approach to the use of CNNs~\cite{Cimpoi2015} for saliency prediction. We, thus, build an image pyramid and compute the CNNs' responses over each scale using the architecture of VGG16~\cite{Simonyan2015} . To combine the convolution responses of each scale, we employ an upsampling procedure and concat the features from the last convolution layer of each scale in order to build a feature map.

Furthermore, we compute the classemes~\cite{Torresani2010} from deep pipeline, that is, the probability of each of the one thousand classes of ImageNet~\cite{Krizhevsky2012} is computed using the fully-connected layers of the VGG16. The classemes are complemented by the low-level scene representation to make the gist of the scene~\cite{Oliva2001}. The classemes and low-level scene features of~\cite{Torralba2006} build a spatial representation of the outside world that is rich enough to convey the meaning of a scene as envisioned in~\cite{Oliva2005}. The feature vector obtained by concatenating classemes and gist features is used for the recognition and retrieval of similar images.

\subsection{Similar image retrieval \& scene bank}

The similar image retrieval unit fetches the information required for building an ensemble of neural predictors from the scene bank. The scene bank holds a set of images in terms of scene representation feature vector, consisting of classemes feature and the gist descriptor, and a neural fixation predictor unit for each image. 

Given the scene representation vector of an input image, denoted as $v^q$, the retrieval method fetches the most $n$ similar images from the set of scene vectors, $\mathbf{V} = \{v_1, \cdots, v_{n^\prime}\}$, using the Euclidean distance, that is, $dist_i = \| v^q - v_i \|$. It then fetches the neural fixation predictor units corresponding to the $n$ images with the smallest $dist_i$ in order to form the ensemble of neural fixation predictors, to be discussed in Section~\ref{sec:spredict}.

\begin{figure}[!t]
\centering
{
\includegraphics[width=1.1in,height=0.8in]{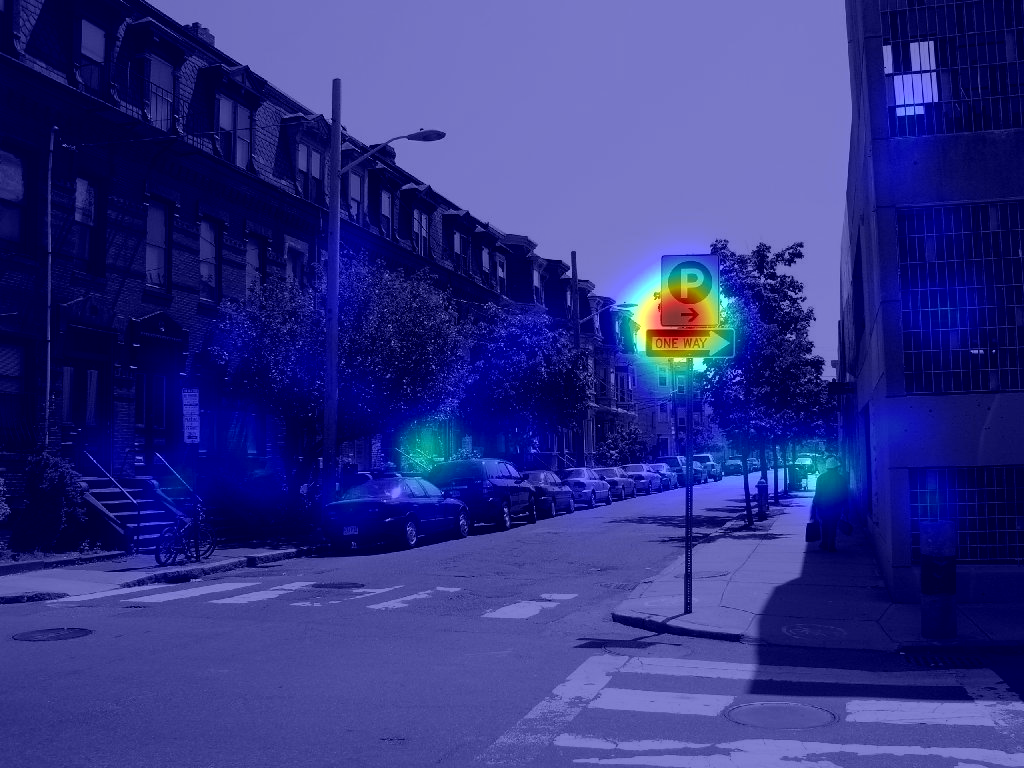} \label{gist_fig_1}}
{
\includegraphics[width=1.1in,height=0.8in]{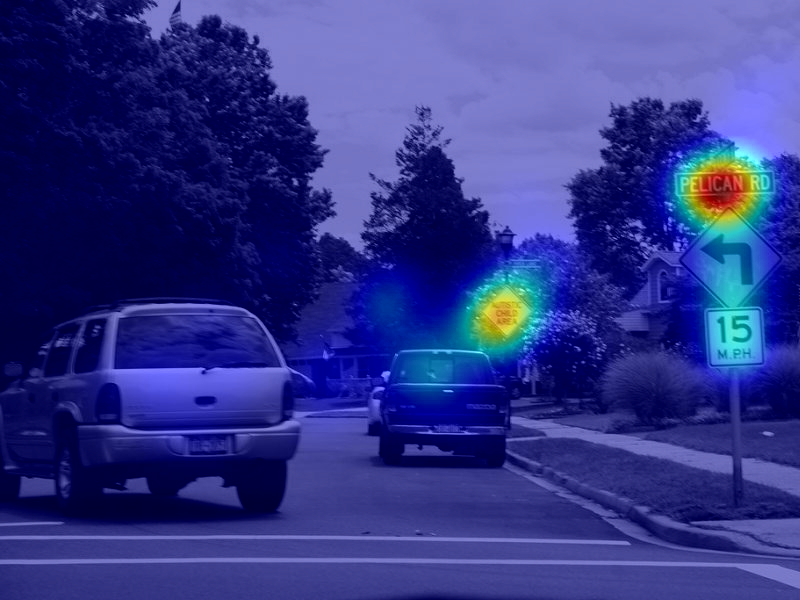} \label{gist_fig_2}} \\
{
\includegraphics[width=1.1in,height=0.8in]{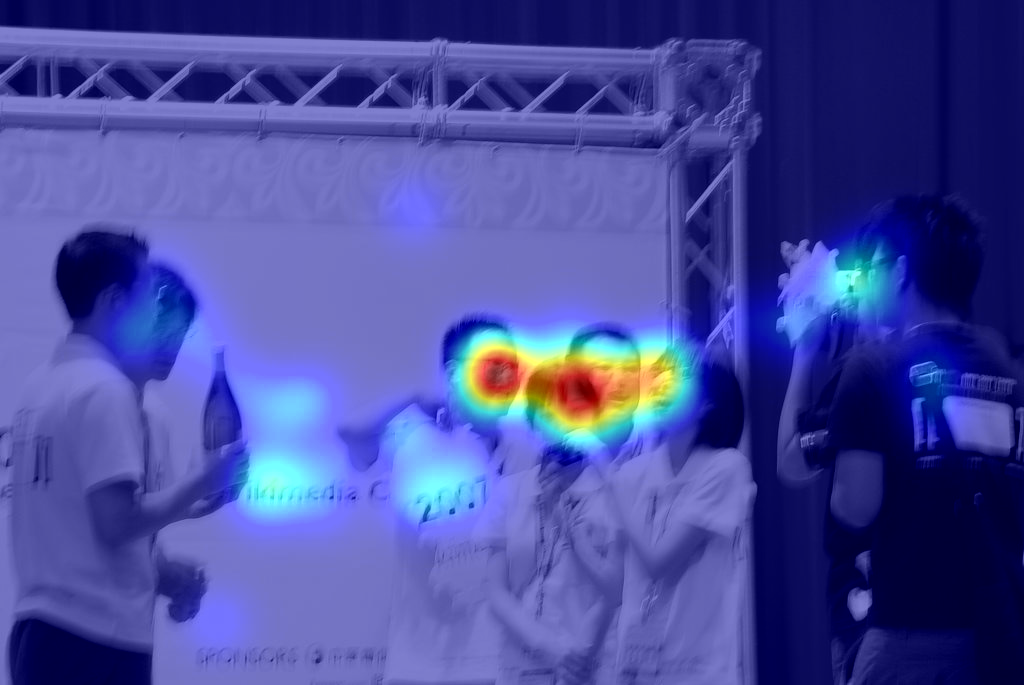} \label{gist_fig_3}} 
{
\includegraphics[width=1.1in,height=0.8in]{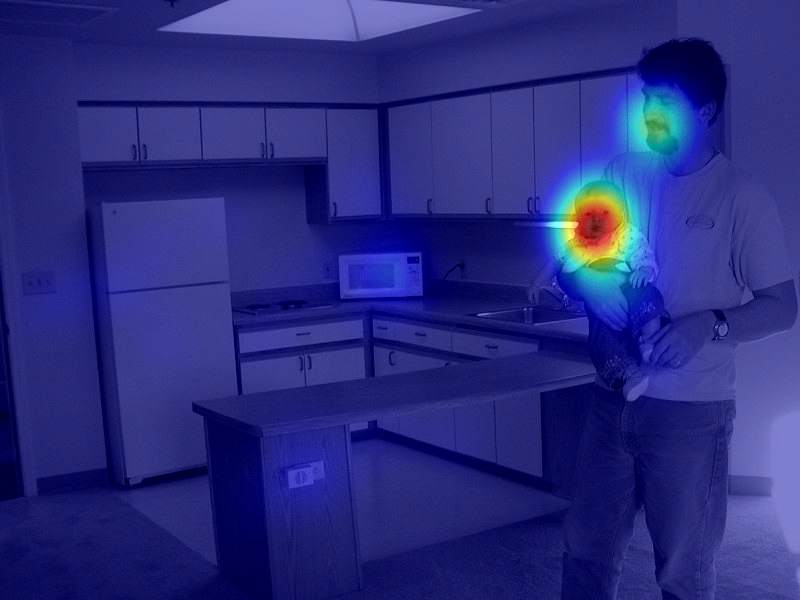} \label{gist_fig_4}} 
\caption{Image retrieval examples. The input (query) image is on the left and its closest match is on the right. The query images are from~\cite{Judd2009} and the closest match is from~\cite{Xu2014}. The observers' fixation density map is overlaid. }
\label{gistFigures}
\end{figure}

Figure~\ref{gistFigures} demonstrates the results of retrieval system. It visualizes a query image and its corresponding most similar retrieved image between two different databases with the observer gaze information overlaid. Interestingly, the retrieved images not only share similar objects and bottom-up structures, but can also have similar attention grabbing regions. It is worth noting that the closest scene is not necessarily of the same scene category, however, it often contains similar low-level and/or high-level perceptual elements.

\subsection{Saliency prediction}
\label{sec:spredict}

We define the saliency of an image in terms of features and locations, that is, $\mathbf{Sal} = p(\mathbf{y}|\mathbf{x},\mathbf{m})$, where $\mathbf{y}$ corresponds to pixel level saliency, $\mathbf{x}$ represents image features and $\mathbf{m}$ is the location. Under the independence assumption, the saliency formulation boils down to the following:

\begin{equation}
\mathbf{Sal} = p(\mathbf{y}|\mathbf{x})p(\mathbf{y}|\mathbf{m}).
\end{equation}

The $p(\mathbf{y}|\mathbf{x})$ corresponds to saliency prediction from image features and $p(\mathbf{y}|\mathbf{m})$ represents a spatial prior. We estimate $p(\mathbf{y}|\mathbf{x})$ using an ensemble of neural predictors and $p(\mathbf{y}|\mathbf{m})$ is learnt from human gaze information.

Figure~\ref{memPipeline} depicts the ensemble of neural saliency predictors.
The ensemble of neural predictors consists of several neural units with equal contributions. In training phase, we train one neural unit for each image in the training set and store them in the scene bank.
In the test phase, the retrieval unit fetches several neural units, corresponding to the $n$ images most similar to the input image.
The ensemble, then, computes the responses of each of the units and aggregates them in order to produce an estimate of $p(\mathbf{y}|\mathbf{x})$, as follows:

\begin{figure}[!t]
\centering
{\label{memPipelineA}
\includegraphics[]{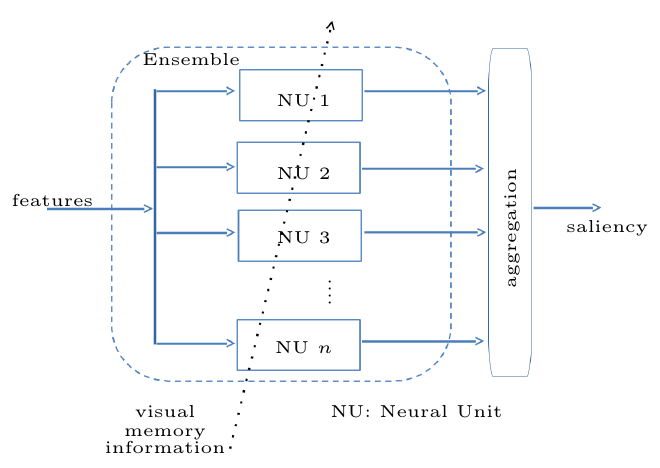} }
\caption{Ensemble of neural saliency predictors.}
\label{memPipeline}
\end{figure}

\begin{equation}
p(\mathbf{y}|\mathbf{x}) = \mathcal{Q} \left( \left( \sum_j \mathcal{C}(\tanh(\mathbf{y}_{j})) \right)^\alpha \right), 
\label{eq:sal1}
\end{equation}

\begin{equation}
\mathcal{C}(x) = \left\lbrace \begin{matrix}
x & x > 0 \\
0		& x \leq 0
\end{matrix}  \right. ,
\label{eq:cfunc}
\end{equation}

\noindent where $\mathcal{Q}(\cdot)$ resizes image or salience data to the size of preference (the size of input image), $\alpha$ is an attenuation factor to emphasize more salient areas, and $\mathbf{y}_{j}$ is the output of the $j$th unit of the ensemble.

\subsubsection{Neural units}

The neural saliency predictor utilizes randomly-weighted single-layer feedforward networks in order to establish a mapping from the feature space to the saliency space. The idea of randomly-weighted single-hidden-layer feedforward networks (SLFNs) can be traced back to the Gamba perceptron~\cite{Minsky1969} followed by others like~\cite{Schmidt1992,Pao1994}. In the neural saliency predictor, we adopt the recent implementation of Extreme Learning Machines (ELM)~\cite{Huang2004}. The theory of ELM  facilitates the implementation of a neural network architecture such that the hidden layer weights can be chosen randomly meanwhile the output layer weights are determined analytically~\cite{Huang2006}. Motivated by better function approximation properties of ELMs~\cite{Huang2006a,Huang2015}, we employ them as the primary entity of the neural saliency prediction.

Having a set of training samples $\{(\mathbf{x}_i, \mathbf{y}_i)\}_{i=1}^N \subset ~\mathbb{R}^k \times ~\mathbb{R}^m$,  the image features $\mathbf{x}_i$ and the corresponding fixation density value $\mathbf{y}_i$ are associated using a SLFNs with $L$ hidden nodes defined as

\begin{equation}
	\mathbf{y}_i = \sum_{j = 1}^{L} \mathbf{\pmb \gamma}_j \mathrm{f}(\mathbf{\pmb \omega}_j \cdot \mathbf{x}_i + b_j),
	\label{eq:fnn}
\end{equation}

\noindent where $\mathrm{f}(\cdot)$ is a nonlinear activation function, $\bold \mathbf{\pmb \gamma}_j \in ~\mathbb{R}^m$ is the output weight vector, 
$\bold \mathbf{\pmb \omega}_j \in \mathbb{R}^k$ is the input weight vector, and $b_j$ is the bias of the $j$th hidden node. The conventional solution to (\ref{eq:fnn}) is gradient-based, which is a slow iterative process that requires to tune all the parameters like $\bold \mathbf{\pmb \gamma}_j$, $\bold  \mathbf{\pmb \omega}_j$ and $b_j$. The iterative scheme is prone to divergence, local minima, and overfitting. The ELM tries to soften such problems and avoid them by random selection of the hidden layer parameters ($\mathbf{\pmb \omega}_j$ and $b_j$) and the estimation of output weights. To this end, (\ref{eq:fnn}) can be rewritten as

\begin{equation}
	\mathbf{Y} = \mathbf{H} \mathbf{\Gamma} ,
	\label{eq:elm}
\end{equation}

\noindent where $\mathbf{Y} = [\mathbf{y}_1~\mathbf{y}_2~ \ldots~\mathbf{y}_N]^T \in \mathbb{R}^{N \times m}$, $\mathbf{\Gamma} = [\bold{\mathbf{\pmb
\gamma}}_1 \bold{~\mathbf{\pmb \gamma}}_2~\ldots~\bold{\mathbf{\pmb \gamma}}_L]^T \in \mathbb{R}^{L \times m}$, and

\begin{equation}
\mathbf{H} = 
	\begin{bmatrix}
		 \mathrm{f}(\mathbf{\pmb \omega}_1 \cdot \mathbf{x}_1 + b_1) & \cdots &  \mathrm{f}(\mathbf{\pmb \omega}_L \cdot  \mathbf{x}_1 + b_L) \\
		 \vdots & \ddots & \vdots \\
		 \mathrm{f}(\mathbf{\pmb \omega}_1  \cdot \mathbf{x}_N + b_1) & \cdots &  \mathrm{f}(\mathbf{\pmb \omega}_L \cdot \mathbf{x}_N + b_L) \\		 
	\end{bmatrix}_{N \times L},
\end{equation}

\noindent which is the hidden layer matrix of the neural network. Once the matrix $\mathbf{H}$ is decided by random selection of input weights and biases, the solution of (\ref{eq:elm}) can be approximated as $\mathbf{\Gamma} = \mathbf{H}^\dagger \mathbf{Y}$, where $\mathbf{H}^\dagger$ is the \emph{Moore-Penrose pseudoinverse} of matrix $\mathbf{H}$.

\subsubsection{Learning spatial prior}

\begin{figure}[!th]
\centering
\includegraphics[scale=0.15]{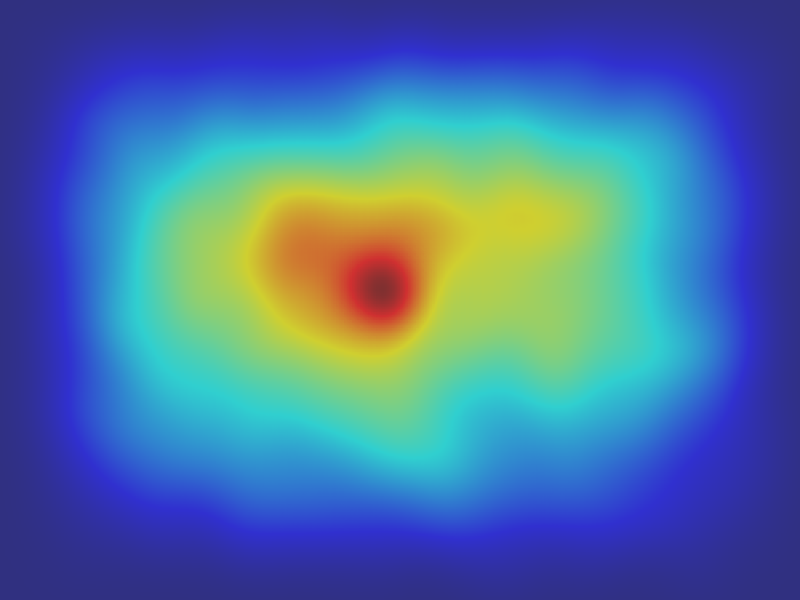}
\caption{Spatial prior learnt from~\cite{Xu2014}.}
\label{fig:spprior}
\end{figure}

In order to learn the spatial prior, $p(\mathbf{y}|\mathbf{m})$, we fit a mixture of Gaussian over the eye fixation data.  We learn the spatial prior using the gaze data of~\cite{Xu2014}, where the number of kernels corresponds to the number of fixation points. The spatial prior puts more weight on the regions that are more agreed by observers. As demonstrated in many saliency research papers, the spatial prior introduces a center-bias effect~\cite{Tatler2009}. The same phenomenon is observed in Figure~\ref{fig:spprior}, depicting the spatial prior.  While there exist arguments on getting advantage of location priors, we address the issue by selecting proper evaluation metrics and benchmarks. It is also worth noting that we are not using summation prior integration, which generally boosts all the regions in the center of the image equally.

\section{Experiments}

We conduct several experiments in order to evaluate the model. The test databases include MIT~\cite{Judd2009}, MIT300~\cite{mit-saliency-benchmark}, and CAT2000~\cite{Borji2015}. The MIT database consists of 1003 images of indoor and outdoor scenes with eye movements of 15 observers. MIT300 consists of 300 natural indoor and outdoor scenes and CAT2000 consists of 4000 images divided into two sets of train and test, with 2000 images in each set. CAT2000 includes 20 categories of images, including, action, affective, art, black \& white, cartoon, fractal, indoor, outdoor, inverted, jumbled, line drawings, low resolution, noisy, object, outdoor man made, outdoor natural, pattern, random, satellite, sketch, and social.

MIT300 and CAT2000 (test set) do not allow the ground-truth access in order to provide a fair comparison. At the moment, they are the widely accepted benchmarks and the results presented are provided by the MIT saliency benchmark team using our submitted maps. The results of the proposed model are also accessible on the benchmark website\footnote{MIT saliency benchmark website:\url{http://saliency.mit.edu/results_mit300.html}} under the acronym ``iSEEL''. 

We learn two ensembles, \emph{ensemble$_{OSIE}$} and \emph{ensemble$_{CAT2k}$}. The first is trained on the OSIE database~\cite{Xu2014} and the latter is trained using the training set of CAT2000. We employ ensemble$_{CAT2k}$ in predicting the CAT2000 test images. The system parameters are optimized for each ensemble. 

In this section, we first explain the system parameters. We then evaluate the performance generalization of the proposed model in comparison with a baseline model using the MIT database. We continue with the Benchmark results on the MIT300 and the CAT2000 databases.

\subsection{System parameters}

The system parameters are the number of neural units in each ensemble, denoted $n$, the number of hidden layers in each unit, $L$, and the attenuation factor,$\alpha$. We furthermore learn a post processing smoothing Gaussian kernel, denoted as $\sigma$, which is used to smooth the model's maps. All the parameters, except the number of hidden nodes are learnt. For each of the ensembles, the number of hidden nodes of each neural unit is fixed and equal to 20. The rest of the parameters of the system are optimized on Toronto database~\cite{Bruce2009}. The tuning cost function minimizes the KL-divergence between the maps of the model and the ground-truth fixation density maps. 

\begin{figure}[!th]
\centering
\subfloat[Ensemble$_{OSIE}$]{
\includegraphics[scale=0.35]{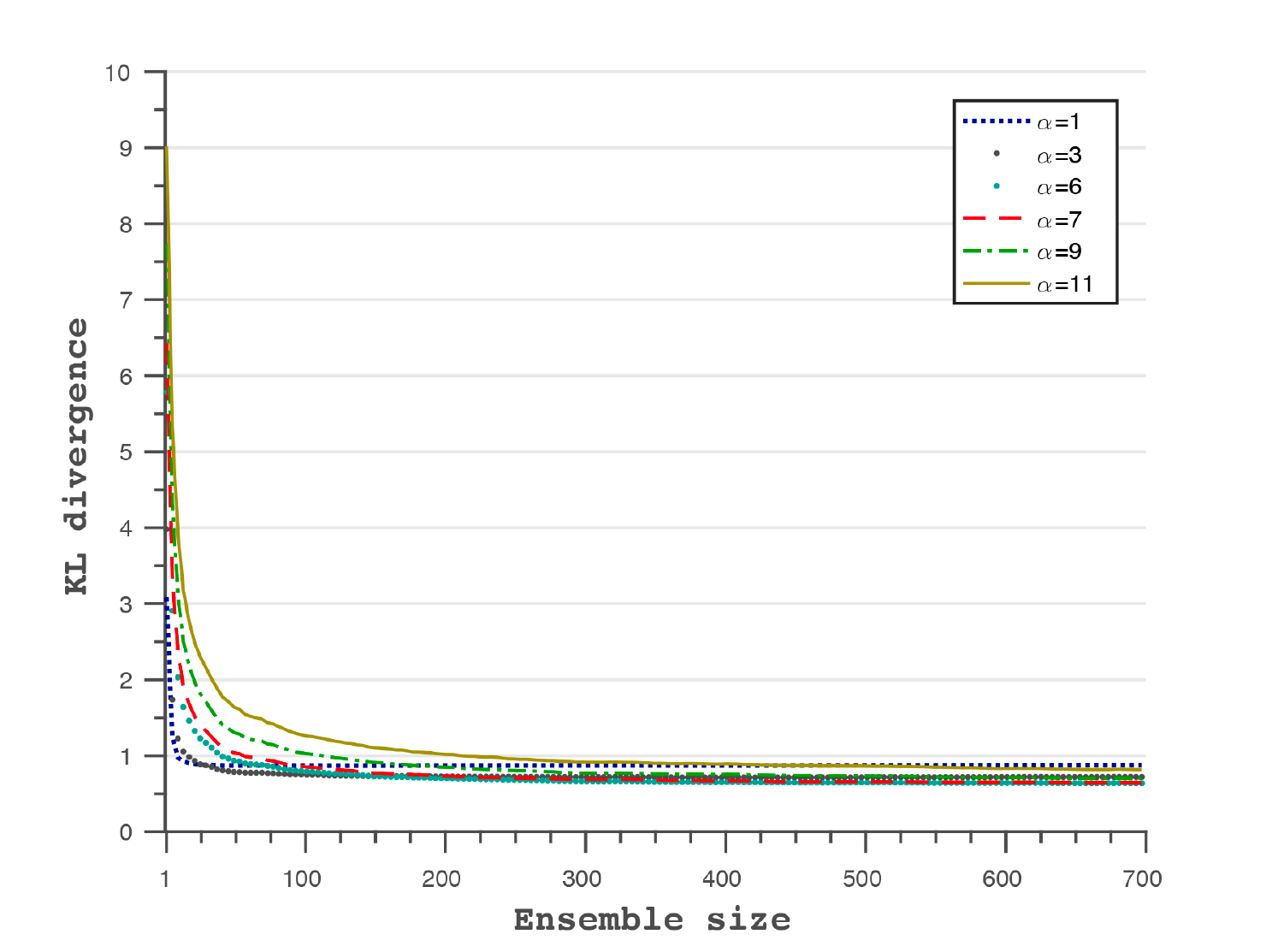}}
\subfloat[Ensemble$_{CAT2k}$]{
\includegraphics[scale=0.35]{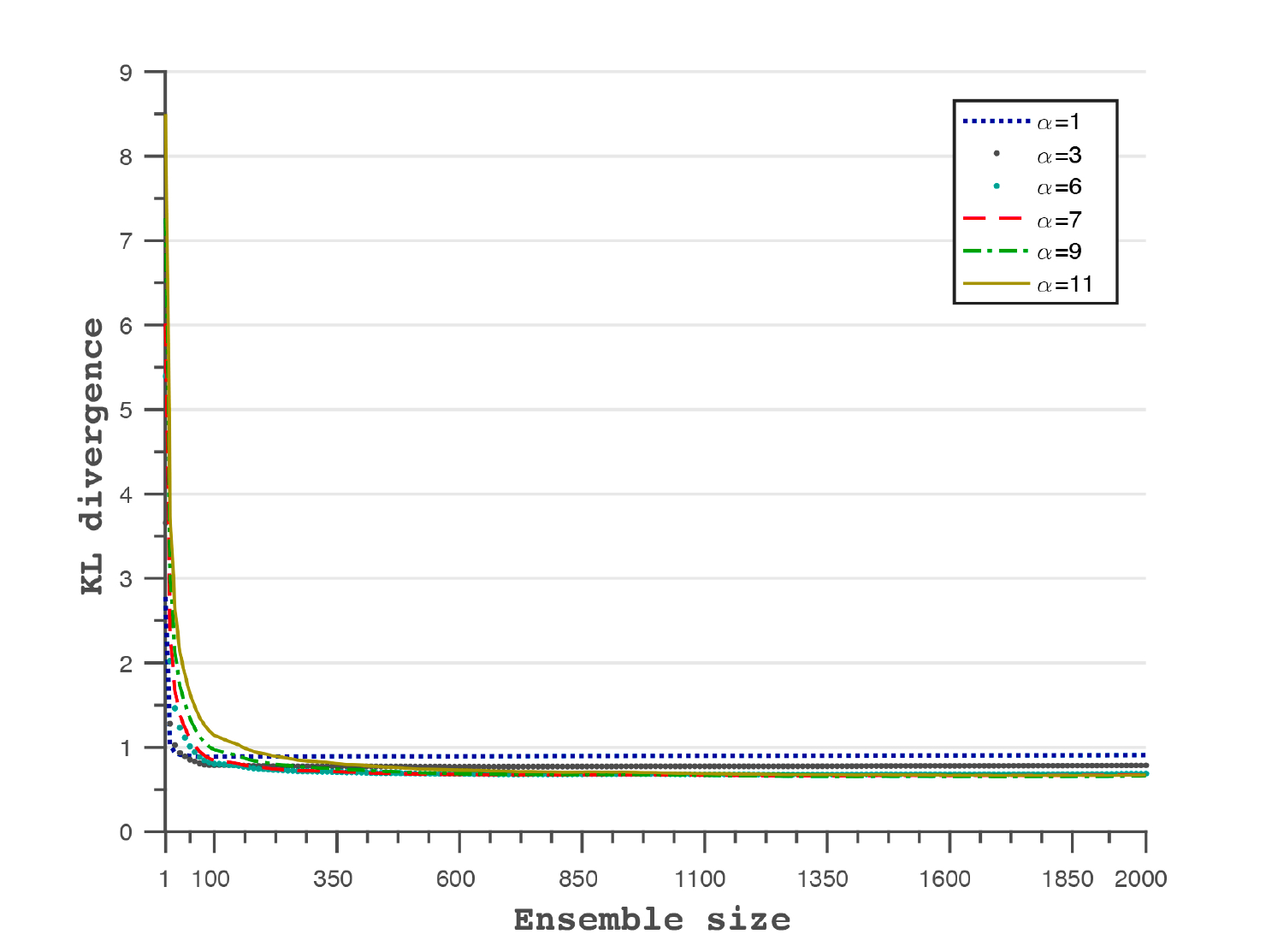}}
\caption{System parameters: the effect of ensemble size and $\alpha$ on the saliency model.}
\label{fig:syspar}
\end{figure}

Figure~\ref{fig:syspar} depicts the effect of the number of neural units in conjunction with the value of the attenuation factor $\alpha$ on the ensemble performance. Based on our observations, an ensemble of size 10 is required to obtain an acceptable result. The optimization of parameters, however, recommend the following parameters for each ensemble, ensemble$_{OSIE}$ : $[n=697, \alpha=6, \sigma=13]$ and  ensemble$_{CAT2k}$: $[n=1710, \alpha=9, \sigma=13]$, where $L=20$ has been fixed.

\subsection{Performance generalization}

To test the generalization of the model, we evaluate its performance using the MIT database~\cite{Judd2009}. We choose the ensemble of deep neural networks (eDN)~\cite{Vig2014} as a baseline model because of the use of deep features and SVM classifiers. The proposed model, however, utilizes an ensemble of ELM regression units. We also evaluate several models including,  AIM~\cite{Bruce2006}, GBVS~\cite{Harel2007}, AWS~\cite{Garcia-Diaz2009}, Judd~\cite{Judd2009}, and FES~\cite{RezazadeganTavakoli2011} for the sake of comparison with traditional models. 

In order to ease the interpretation of evaluation, we choose a subset of scores that complement each other. We employ shuffled AUC (sAUC, an AUC metric that is robust towards center bias), similarity metric (SIM, a metric indicating how two distributions resemble each other~\cite{Judd2012}), and normalized scanpath saliency (NSS, a metric to measure consistency with human fixation locations). NSS and sAUC scores are utilized in~\cite{Borji2013}, which we borrow part of the scores from, and complement them with the SIM score.

\begin{figure}[!th]
\centering
\subfloat[]{
\includegraphics[scale=0.25]{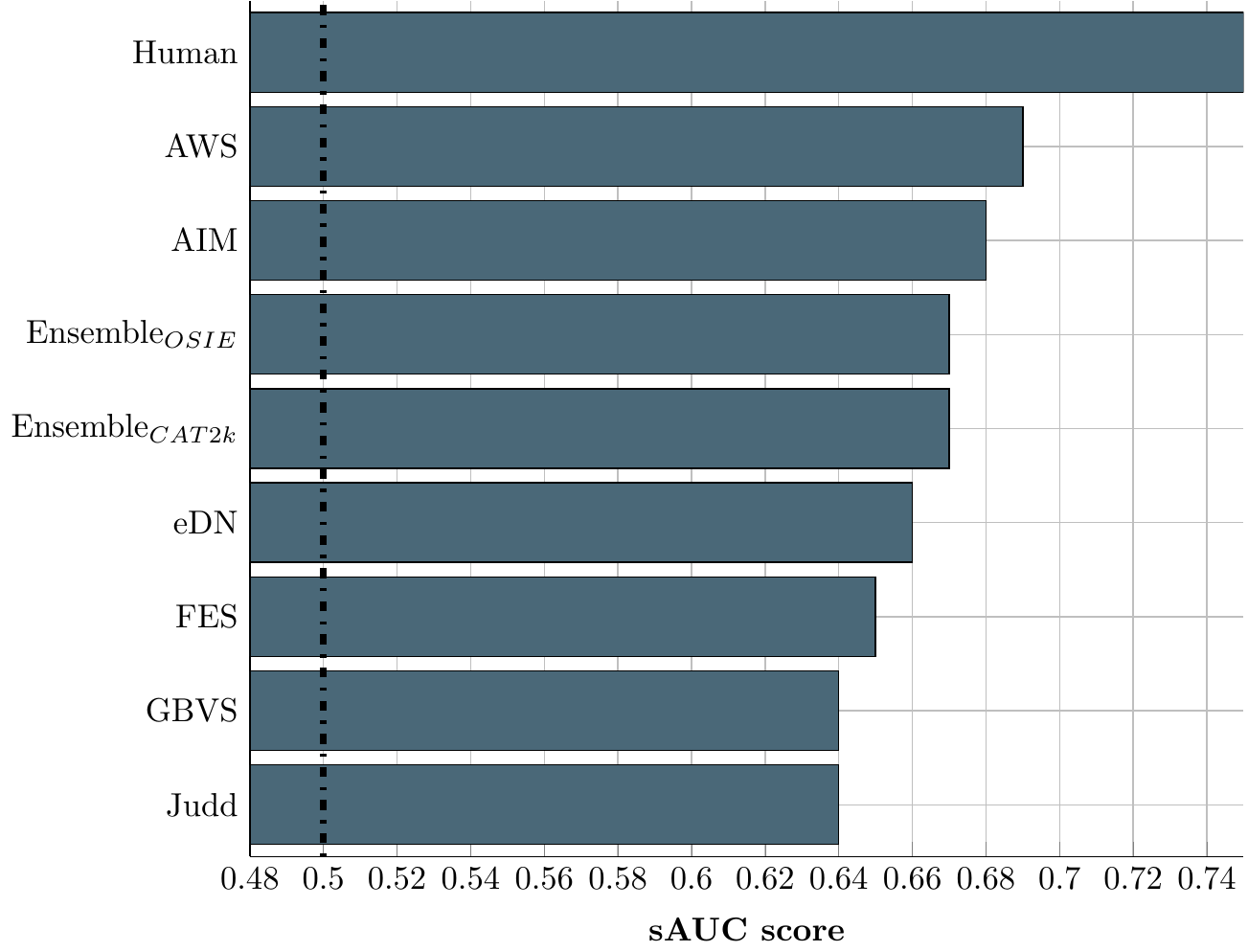}}
\subfloat[]{
\includegraphics[scale=0.25]{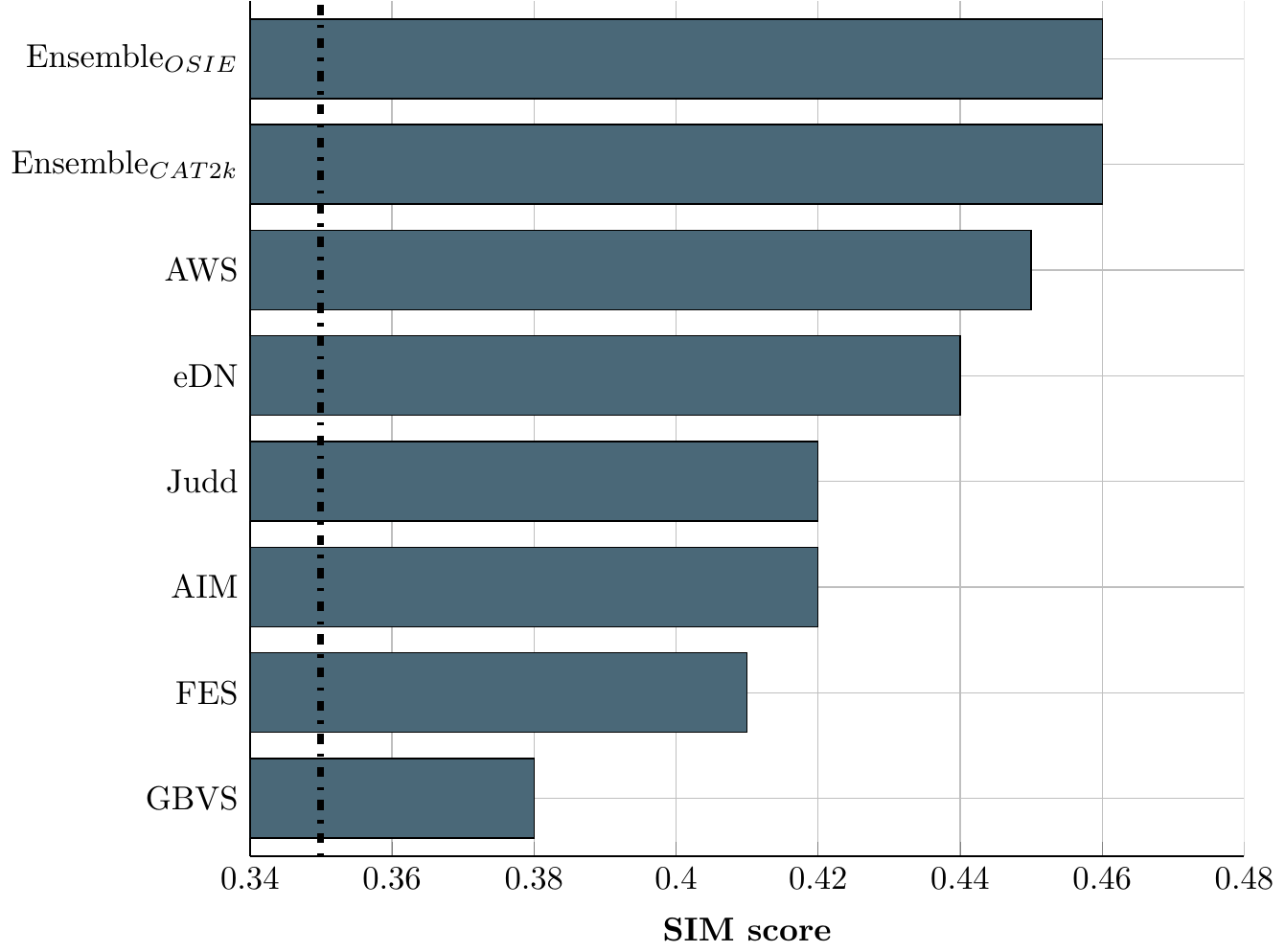}}
\subfloat[]{
\includegraphics[scale=0.25]{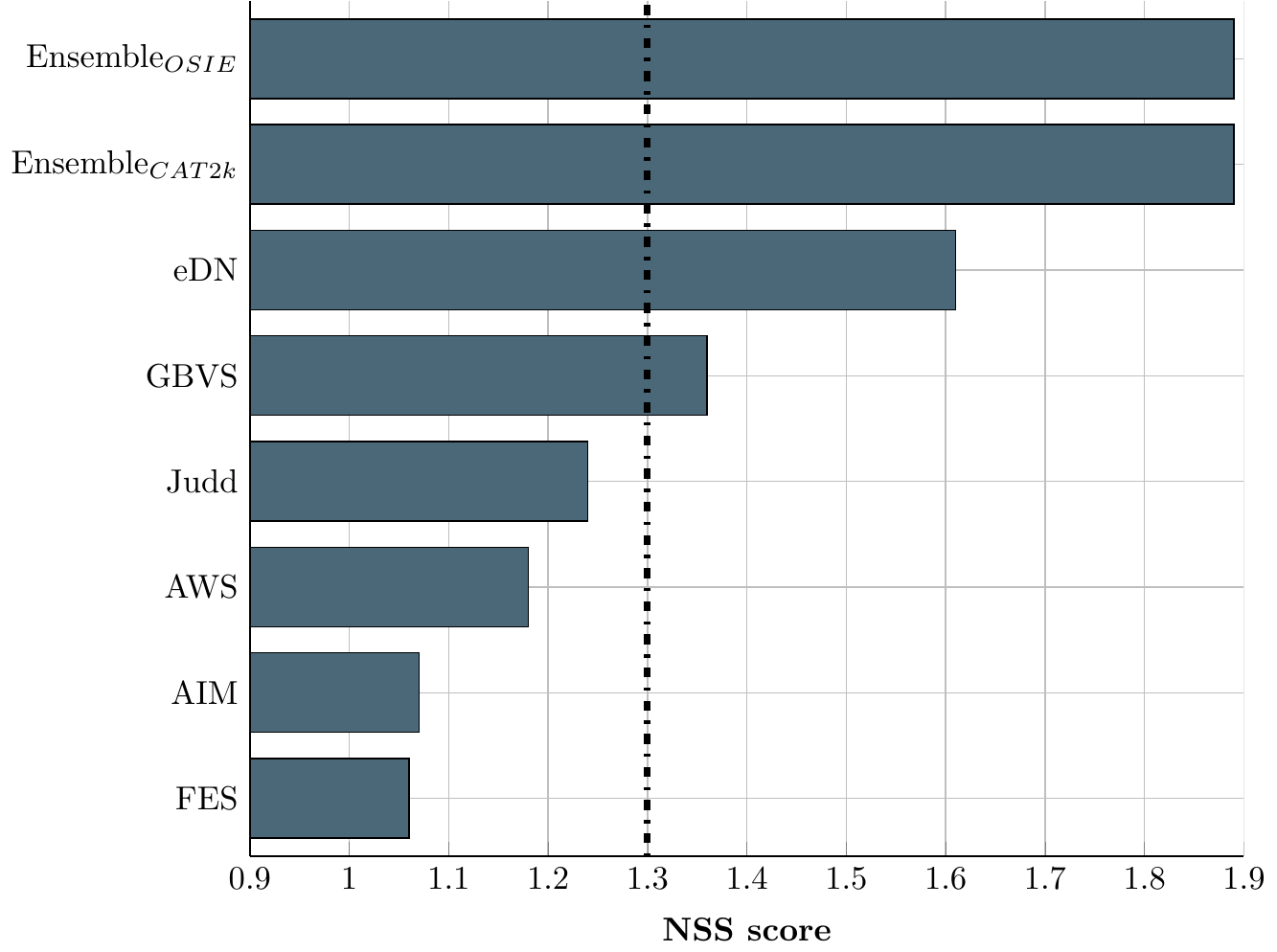}}
\caption{Performance generalization: the performance of the proposed model compared to traditional models and eDN~\cite{Vig2014} as a baseline model. The dashed vertical line indicates the performance of a Gaussian dummy model. The human score for SIM and NSS are 1 and 3.1, respectively.}
\label{fig:perfgen}
\end{figure}

Figure~\ref{fig:perfgen} reports the results. %In this experiment, Ensemble$_{OSIE}$ is used because the nature of OSIE and MIT1003 is similar, both consist of indoor and outdoor images.
As depicted, the proposed model outperforms all other models on two metrics and outperforms the eDN on all the three metrics. The highest gain compared to the eDN is on the NSS score, indicating a high consistency with human fixation locations which explains the high SIM score as well. To summarize, the proposed model generalizes well and has the edge over traditional models. We later compare the proposed model with the recent state-of-the-art models on well-established benchmarks. 

\subsection{Benchmark}
 
Many of the recent deep saliency models have their codes and maps unavailable to public, making comparisons difficult. We, hence, rely on available benchmarks.
We report the performance using all the metrics and published works, reported on the MIT benchmark. For brevity, the focus will be on recent top-performing models. The results also include the performance of {``Infinite Human''} and {``Mean One Human''} to indicate how well a model performs in comparison with mean eye position of several human (upper-bound performance) and the on average performance of one human, respectively.

\paragraph{Results on MIT300} Table~\ref{table:MIT300} summarizes the performance comparison, where the proposed model is 4th among published works on this benchmark on the basis of NSS. MIT300 is the largest benchmark with over 60 models at the time of this writing. We, however, report the best performing models and the most recent state-of-the-art ones. The comparison indicates that the models are becoming powerful enough to capture fixation location.
It is, hence, difficult to distinguish them from each other on many metrics. NSS, however, seems to be the most informative metric that determines the models' performance well, particularly for top-performing models that judging AUC-based metrics and Similarity-based metrics are difficult.

\begin{table}
\tiny
\renewcommand{\arraystretch}{1.3}
% if using array.sty, it might be a good idea to tweak the value of
% \extrarowheight as needed to properly center the text within the cells
\caption{MIT300 Benchmark results, sorted using NSS.}
\label{table:MIT300}
\centering

% Some packages, such as MDW tools, offer better commands for making tables
% than the plain LaTeX2e tabular which is used here.
\begin{tabular}{l|c|c|c|c|c|c|c|c}
\multirow{2}{*}{\bfseries Model} &   \multicolumn{3}{c|}{\bfseries AUC-based metrics} & \multicolumn{4}{c|}{\bfseries Similarty-based metrics} & \\
\cline{2-8}
	& Judd & Borji & shuffled & SIM & EMD & CC & KL  & NSS  \\
\hline
\hline
\bfseries Infinite Human & 0.92 & 0.88 & 0.81 & 1.00 & 0 & 1 & 0 & 3.29\\
\bfseries SALICON~\cite{Huang2015b} & 0.87 & 0.85 & 0.74 & 0.60 & 2.62 & 0.74 & 0.54 & 2.12\\
\bfseries PDP~\cite{Jetley2016} & 0.85 & 0.80 & 0.73 & 0.60 & 2.58& 0.70 & 0.92 & 2.05 \\
\bfseries ML-Net~\cite{Cornia2016} & 0.85 & 0.75 & 0.70 & 0.59 & 2.63 & 0.67 & 1.10 & 2.05\\
\bfseries ensemble$_{OSIE}$(iSEEL) & 0.84 & 0.81 & 0.68 & 0.57 & 2.72 & 0.65 & 0.65 & 1.78 \\
\bfseries Mean One Human  & 0.80 & 0.66 & 0.63 & 0.38 & 3.48 & 0.52 & 6.19 & 1.65\\
\bfseries SalNet~\cite{Pan2016} & 0.83 & 0.82 & 0.69 & 0.52 & 3.31 & 0.58 & 0.81 & 1.51\\
\bfseries BMS~\cite{Zhang2013} & 0.83 & 0.82 & 0.65 & 0.51 & 3.35 & 0.55 & 0.81 & 1.41 \\
\bfseries Mr-CNN~\cite{Liu2015} & 0.79 & 0.75 & 0.69 & 0.48 & 3.71 & 0.48 & 1.08 & 1.37 \\
\bfseries eDN~\cite{Vig2014} & 0.82 & 0.81 & 0.62 & 0.41 & 4.56 & 0.45 & 1.14 & 1.14\\
\end{tabular}
\end{table}

\paragraph{Results on CAT2000} Table~\ref{table:CAT2K} contains the performance comparison on the CAT2000 database. 19 models, which are mostly traditional ones, are evaluated on this database. The proposed model, ensemble$_{CAT2k}$, ranks similarly with BMS~\cite{Zhang2013} at the top of the ranking. Both models produce the highest NSS score among models and on average have indistinguishable values for the AUC-based and the Similarity-based metrics.

\begin{table}
\tiny
\renewcommand{\arraystretch}{1.3}
% if using array.sty, it might be a good idea to tweak the value of
% \extrarowheight as needed to properly center the text within the cells
\caption{CAT2000 Benchmark results, sorted using NSS.}
\label{table:CAT2K}
\centering

% Some packages, such as MDW tools, offer better commands for making tables
% than the plain LaTeX2e tabular which is used here.
\begin{tabular}{l|c|c|c|c|c|c|c|c}
\multirow{2}{*}{\bfseries Model} &   \multicolumn{3}{c|}{\bfseries AUC-based metrics} & \multicolumn{4}{c|}{\bfseries Similarty-based metrics} & \\
\cline{2-8}
	& Judd & Borji & shuffled & SIM & EMD & CC & KL  & NSS  \\
\hline
\hline
\bfseries Infinite Human & 0.90 & 0.84 & 0.62 & 1.00 & 0 & 1 & 0 & 2.85\\
\bfseries ensemble$_{CAT2k}$(iSEEL) & 0.84 & 0.81 & 0.59 & 0.62 & 1.78 & 0.66 & 0.92 & 1.67 \\
\bfseries BMS~\cite{Zhang2013} & 0.85 & 0.84 & 0.59 & 0.61 & 1.95 & 0.67 & 0.83 & 1.67 \\
%\bfseries Permutation Control~\cite{Koehler2014} & 0.80 & 0.71 & 0.50 & 0.55 & 2.25 & 0.63 & 2.42 & 1.63  \\
\bfseries ensemble$_{OSIE}$ & 0.83 & 0.81 & 0.59 & 0.59 & 2.24 & 0.64 & 0.67 & 1.62 \\
\bfseries FES~\cite{Zhang2013} & 0.82 & 0.76 & 0.54 & 0.57 & 2.24 & 0.64 & 2.10 & 1.61 \\
\bfseries Mean One Human  & 0.76 & 0.67 & 0.56 & 0.43 & 2.51 & 0.56 & 7.77 & 1.54\\
\bfseries Judd~\cite{Judd2009} & 0.84 & 0.84 & 0.56 & 0.46 & 3.60 & 0.54 & 0.94 & 1.30\\
\bfseries eDN~\cite{Vig2014} & 0.85 & 0.84 & 0.55 & 0.52 & 2.64 & 0.54 & 0.97 & 1.30\\
\end{tabular}
\end{table}

We also evaluate ensemble$_{OSIE}$ along with ensemble$_{CAT2k}$ in order to further investigate the improvements caused by incorporating similar images in the training phase. Backing the hypothesis, the ensemble trained on CAT2000 outperforms the ensemble that is learnt from only indoor and outdoor images of OSIE in terms of the overall scores. 

\begin{figure}[!th]
\centering
%\subfloat[]{
\includegraphics[scale=0.25]{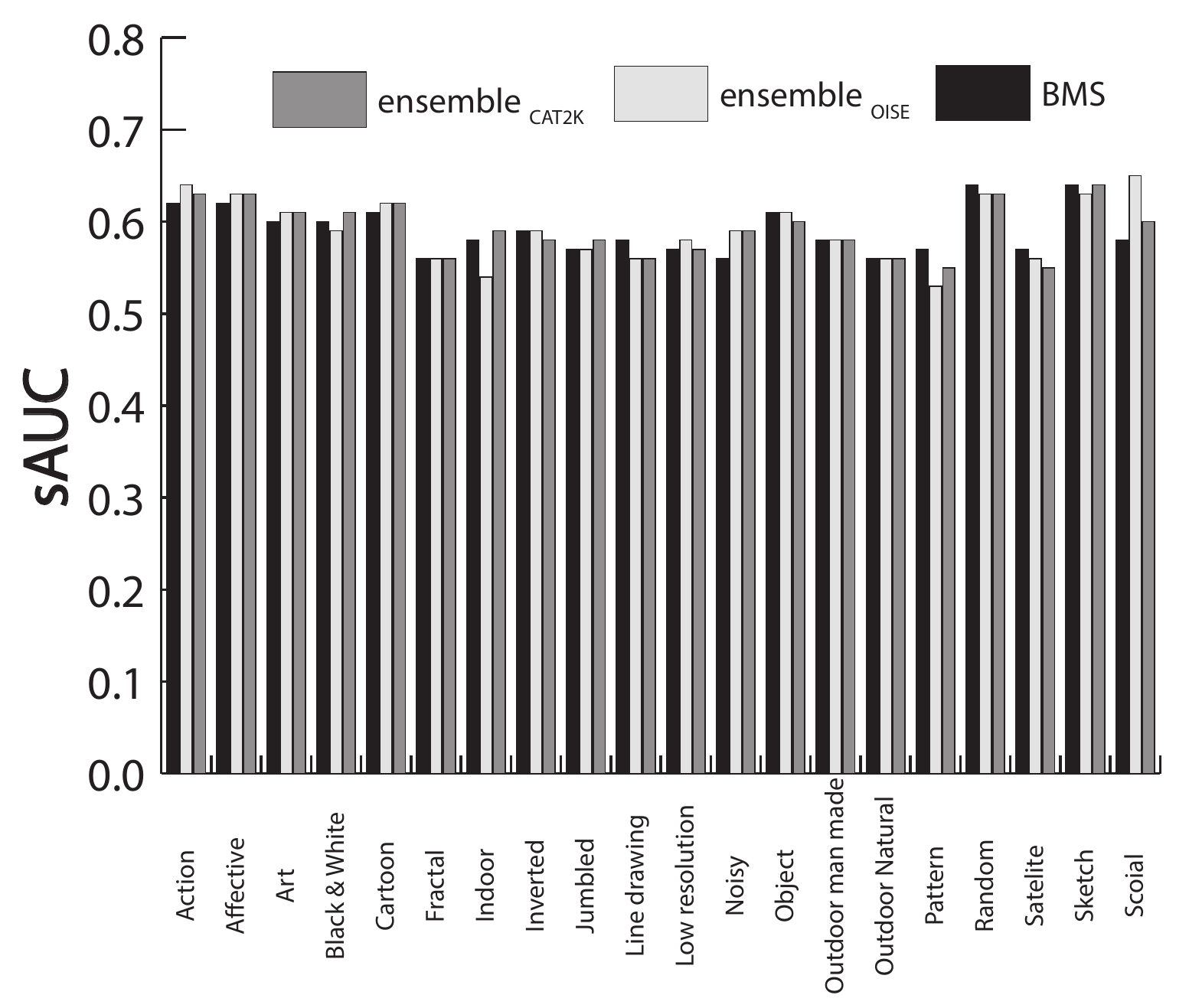}
%}
%\subfloat[]{
\includegraphics[scale=0.25]{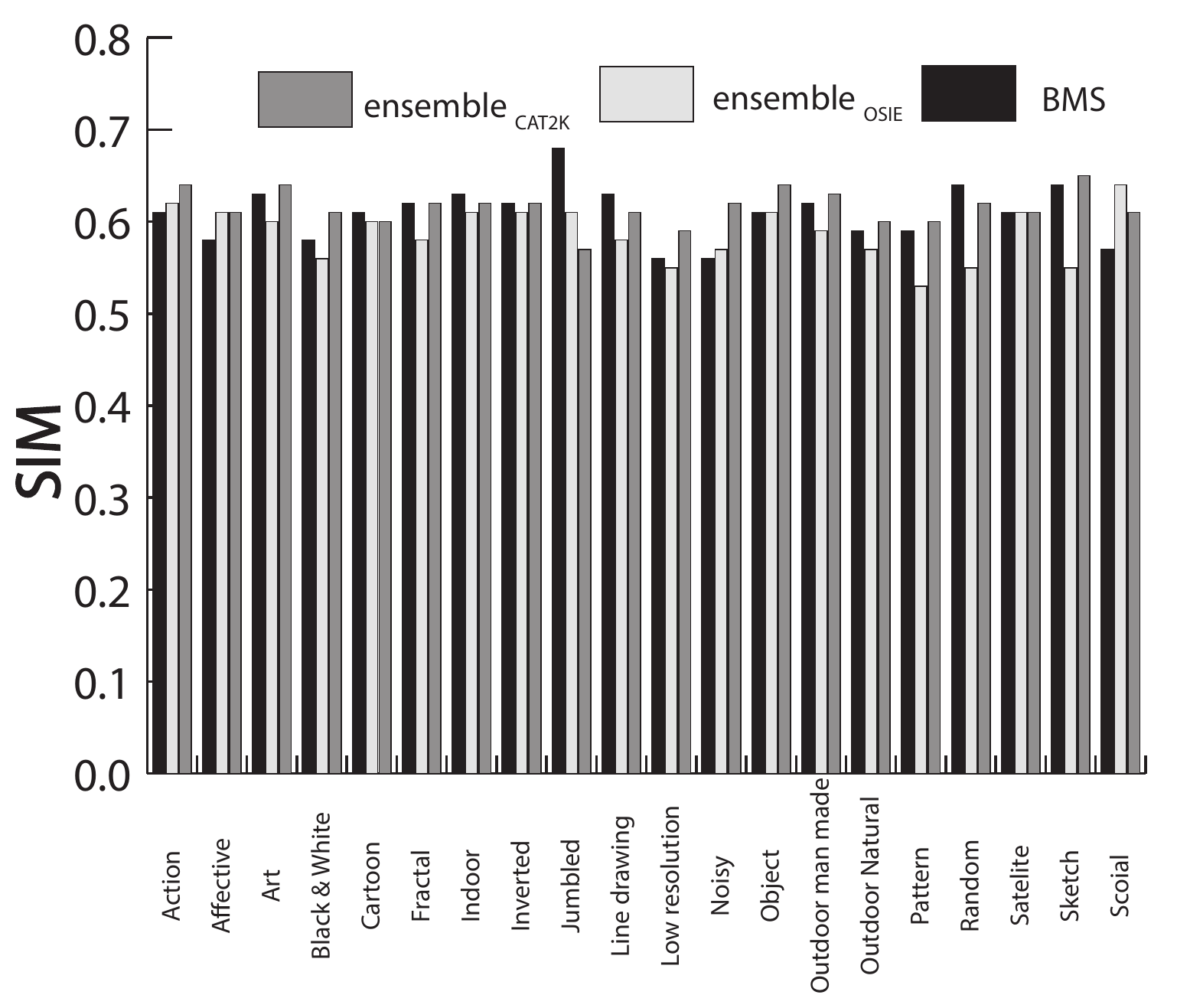}
%}
%\subfloat[]{
\includegraphics[scale=0.25]{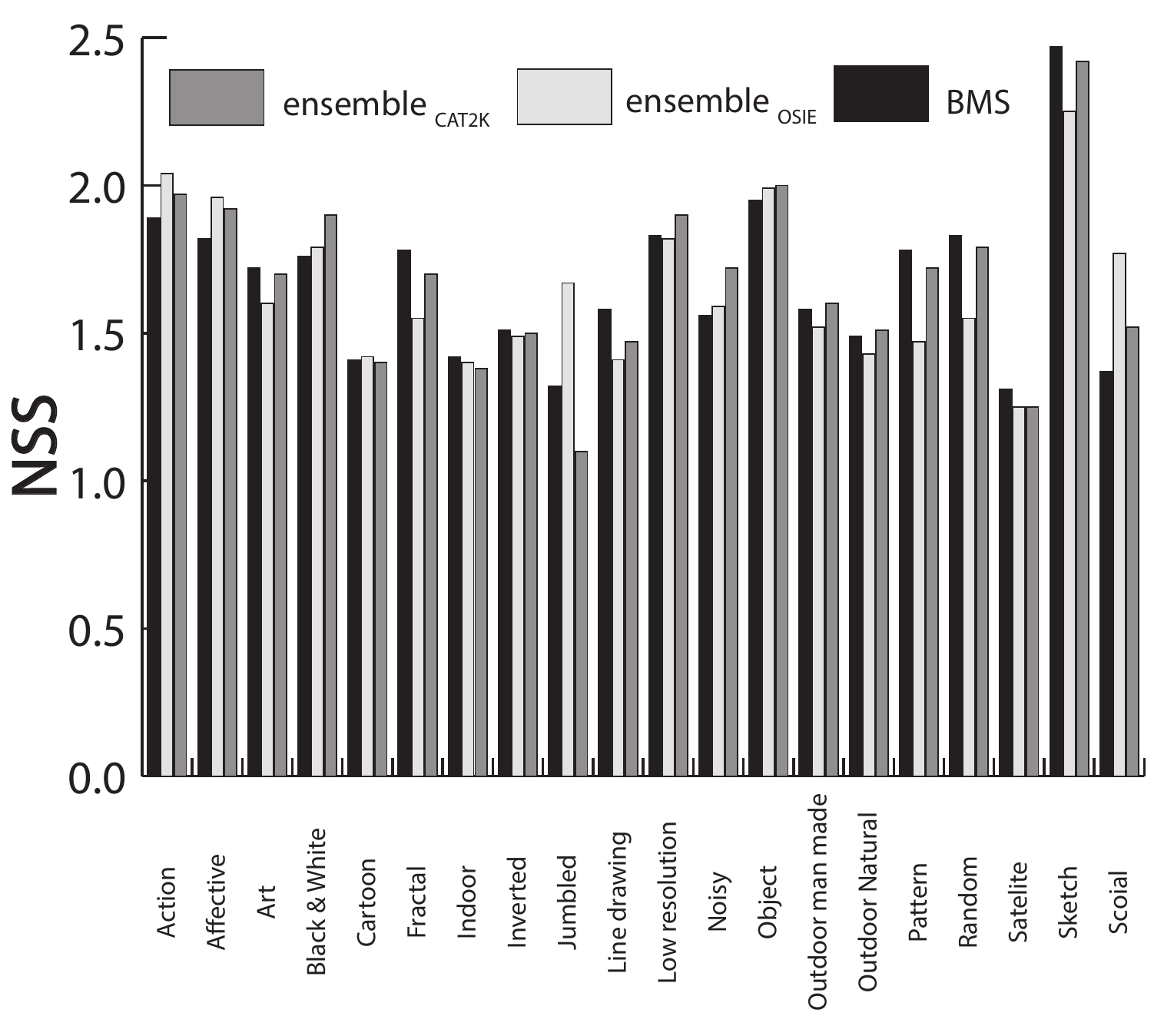}
%}
\caption{Performance on categories of CAT2000: the performance of the proposed model compared to BMS.}
\label{fig:perfCategory}
\end{figure}

We look into the performance of the models in each of the twenty class categories of CAT2000 database. To be concise, we investigate ensemble$_{CAT2k}$, ensemble$_{OSIE}$, and BMS, which are the top three best performing models, using the three metrics of shuffled AUC (sAUC), SIM, and NSS. The results are summarized in Figure~\ref{fig:perfCategory}. The proposed model, both ensemble$_{CAT2k}$ and ensemble$_{OSIE}$, are outperforming the BMS on low resolution, noisy, outdoor, black \& white, action, affective and social categories. 
The BMS seems performing better when there is no particular contextual information and more low-level feature interactions matter, e.g., fractal category, and pattern. The other categories are, however, more difficult to judge. Overall, it seems the three models can complement each other in the areas where one falls behind the others.

\section{Discussion \& conclusion}

We demonstrated the usefulness of scene similarity in predicting the saliency motivated by the effect of the familiarity of a scene on the observer's eye movements. The idea can, however, be easily extended to the utilization of observers' eye movements in task-specific models, where a model is trained for a specific task and experts' eye movements are incorporated. An expert approach for solving a specific task is different from that of a naive observer. Thus, we can consider the encoding of expert observers' eye movements as an implicit expert knowledge utilization, which can be handy in scenarios of scene analysis such as spotting object-specific anomalies from saliency maps in order to reduce the search time.

We introduced a saliency model with the motive of exploiting the effect of immediate scene recall on the human perception.
The proposed model uses randomly-weighted neural networks as an ensemble architecture.
It establishes a mapping from a feature space, consisting of deep features, to the saliency space. 
The saliency prediction relies only on the neural units corresponding to the images that are similar to the input image.
The neural units are pre-trained and stored in a scene bank from a handful of images. For each neural unit, the scene bank also stores a scene descriptor, consisting of classemes and gist descriptor. To find the similar images from scene bank, the proposed model employs the distance between the scene descriptor of the input image and neural units.

The proposed model was evaluated on several databases. The results were reported on two well-established benchmark databases by the MIT benchmark team, namely MIT300 and CAT2000. Among the published methods and on the basis of NSS, consistency with the locations of human fixation, the proposed method was ranked 4th and 1st (in conjunction with BMS) on MIT300 and CAT2000, respectively. The results indicate benefit from learning saliency from images similar to the input image. The code for the proposed model is available at:~\url{http://github.com/hrtavakoli/iseel}.

\section*{Acknowledgement}
Hamed R.-Tavakoli and Jorma Laaksonen were supported by the Finnish Center of Excellence in Computational Inference Research (COIN). The authors would like to thank the MIT saliency benchmark team, particularly Zoya Bylinskii, for their quick response on benchmark request.

\bibliography{memSalBiblo}

\end{document}